\documentclass[letterpaper, 10 pt, conference]{ieeeconf}  % Comment this line out if you need a4paper

\usepackage[T1]{fontenc}
\usepackage{amsmath,amssymb,amsfonts}
\usepackage{bm}
\usepackage{cite}
\usepackage{microtype}

\usepackage{booktabs}
\usepackage{multirow}
\usepackage{graphicx}

\providecommand{\paperassetpath}{}
\graphicspath{{\paperassetpath}}

\ifdefined\pdfsuppresswarningpagegroup
\fi
\newcommand{\tablefont}{\fontsize{9}{10}\selectfont}
\newcommand{\tablesetup}{%
    \tablefont
    \renewcommand{\arraystretch}{1.00}%
    \setlength{\tabcolsep}{2pt}%
}
\IEEEoverridecommandlockouts                              % This command is only needed if 
\newif\ifanonymous
\anonymousfalse       % camera-ready

\title{\LARGE \bf
Robustness as a Transferable Skill: Cross-Task Adversarial Defense for Vision-Language-Action Policies}

\ifanonymous
\author{Anonymous Author(s)\\
Anonymous Institution\\
{\ttfamily anonymous@anonymous.com}}
\else

\author{
Jinquan Zhang$^{1,2}$,
Run Yang$^{1,2}$,
Dongfu Yin$^{1,*}$,
Qiang Lai$^{1,2}$,\\
Yufeng Yan$^{1,2}$,
Pengshuai Yin$^{1}$,
and F. Richard Yu$^{3}$%
\thanks{This work was supported in part by the Shenzhen Science and Technology Program under Grant KJZD20240903104400001.}%
\thanks{$^{1}$Guangdong Laboratory of Artificial Intelligence and Digital Economy (SZ), Shenzhen, China.}%
\thanks{$^{2}$Shenzhen University, Shenzhen, China.}%
\thanks{$^{3}$Carleton University, Ottawa, Canada.}%
\thanks{$^{*}$Corresponding author: Dongfu Yin (yindongfu@gml.ac.cn).}%
}

\fi

\begin{document}

\maketitle
\thispagestyle{empty}
\pagestyle{empty}
\bstctlcite{BSTcontrol}

%%%%%%%%%%%%%%%%%%%%%%%%%%%%%%%%%%%%%%%%%%%%%%%%%%%%%%%%%%%%%%%%%%%%%%%%%%%%%%%%

\begin{abstract}

Vision-Language-Action (VLA) policies are vulnerable to
adversarial visual perturbations, while defenses often require
task-specific adversarial fine-tuning. We investigate whether
robustness learned on source manipulation tasks can protect
independently specialized target policies without target-side
adversarial training. We first show that even a single source
defense update improves robustness on held-out tasks,
establishing cross-task defense reuse independently of
multi-source aggregation. To combine source updates, we
introduce Source-Norm Transferable Robustness Vector (SN-TRV),
which retains the direction of their mean while calibrating
its magnitude using source-update norms. The resulting
update is applied to the target visual encoder without
target adversarial data, target-side gradient optimization,
or additional inference modules. Evaluations across three
VLA backbones, fixed and adaptive attacks, and different
source defense recipes show improved robustness over naive
averaging and identify limits in attack coverage. Physical
PiPER trials increase average success under printed
adversarial patches from 20.3\% to 45.8\%.

\end{abstract}

%%%%%%%%%%%%%%%%%%%%%%%%%%%%%%%%%%%%%%%%%%%%%%%%%%%%%%%%%%%%%%%%%%%%%%%%%%%%%%%%
\section{Introduction}

Vision-Language-Action (VLA) policies have recently enabled
general-purpose robotic manipulation by integrating visual perception,
language understanding, and low-level control. Models such as RT-2,
OpenVLA, Octo, and $\pi_0$ demonstrate the potential of scaling
vision-language knowledge to robotic tasks~\cite{zitkovich2023rt2,
kim2025openvla,octo2024,black2025pi0}. However, the reliance on visual
observations also exposes VLA policies to adversarial perturbations that can
directly cause physical execution failures.

Recent studies reveal that physically realizable adversarial patches can
severely degrade VLA manipulation performance. EDPA introduces
model-agnostic adversarial patches and shows that adversarial fine-tuning of
the visual encoder can recover robustness against such attacks
~\cite{xu2025edpa}. Other works investigate stronger transferable attacks
and robustness evaluation frameworks, including UPA-RFAS, AttackVLA, and
physical variation benchmarks~\cite{lu2026uparfas,li2025attackvla,
liu2025evavla}. Correspondingly, recent defenses improve robustness through
adversarial fine-tuning, attention-aware optimization, lightweight adapters,
or certified robustness techniques
~\cite{zhang2026sarf,yin2026vlaguard,fu2026stablevla,lu2026certvla}.

\begin{figure}[!t]
    \centering
    \includegraphics[width=\linewidth]{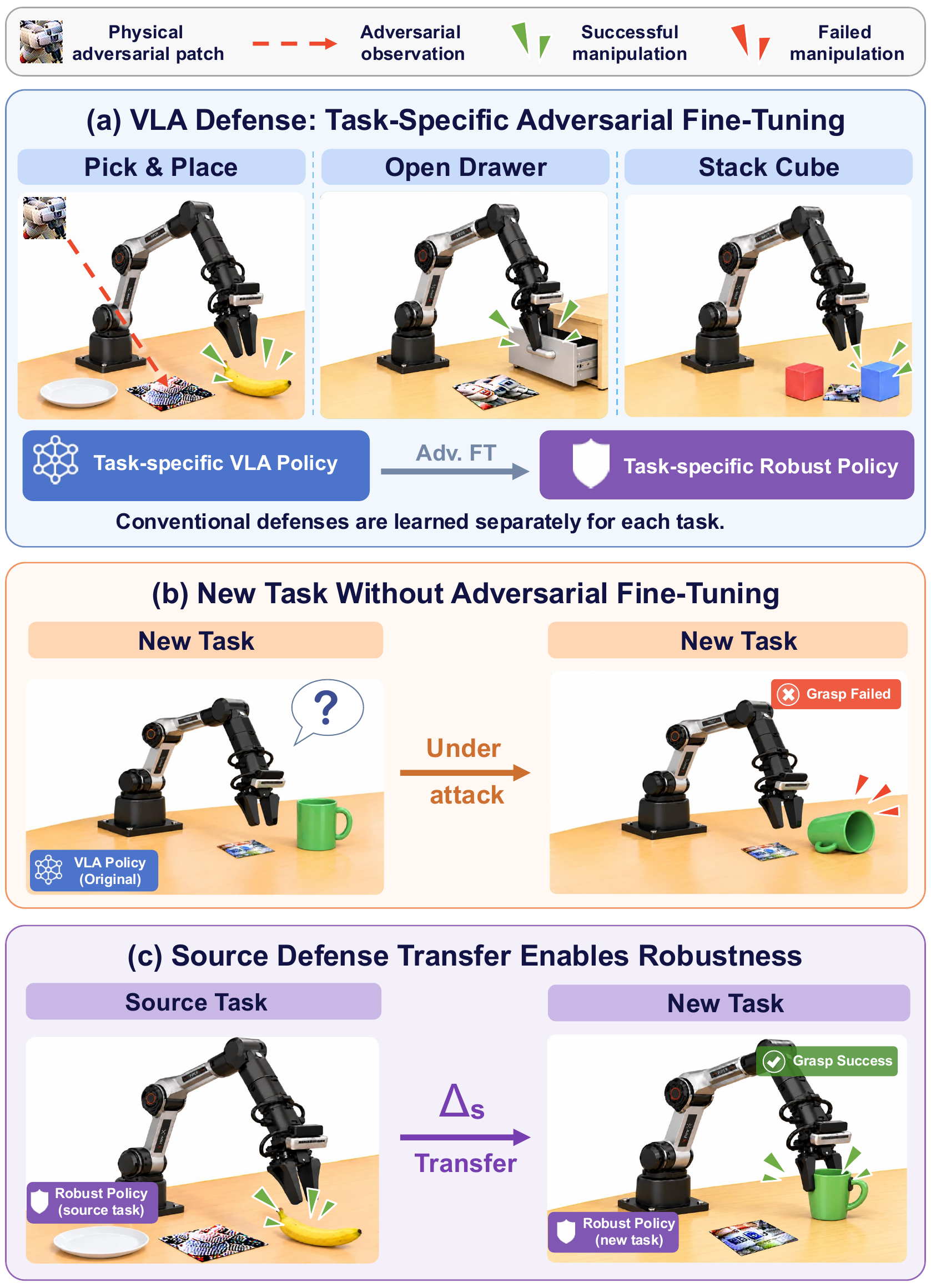}
    \caption{
Cross-task adversarial robustness transfer.
(a) Conventional defenses require task-specific adversarial fine-tuning.
(b) An undefended target policy fails under an adversarial observation.
(c) A source-task visual defense update protects the independently
specialized target policy without target-task adversarial fine-tuning.
}
    \label{fig:motivation}
\end{figure}

Despite these advances, existing VLA defenses
primarily acquire robustness by optimizing the
policy to be protected.
When a policy is independently specialized for
a new manipulation task, adversarial fine-tuning
may need to be repeated even if robust source
policies are already available.
This motivates the following question:

\emph{Can adversarial robustness learned on one manipulation task be reused
as a transferable capability to protect another independently specialized
policy?}

We find that visual defense updates learned on
source manipulation suites can substantially
improve the adversarial robustness of independently
specialized target policies.
Importantly, this effect is observed even when
transferring a single source update, establishing
defense reuse independently of any multi-source
aggregation method.

Combining multiple source defenses, however,
raises two questions: how to estimate a
transferable defense direction and how to
determine its update magnitude.
Naive averaging provides a natural aggregate direction, but partially misaligned source updates can attenuate its magnitude.

To address this issue, we propose
\emph{Source-Norm Transferable Robustness Vector (SN-TRV)}.
SN-TRV estimates a source-consensus defense direction and recomposes
it with a reference magnitude derived from source defense updates.
The resulting vector can be transferred to a held-out target policy
without target adversarial examples, target-side gradient optimization,
or additional inference-time modules.

The contributions of this work are:

\begin{itemize}
    \item We establish empirical evidence of cross-task
    adversarial defense reuse in VLA policies, showing
    that even a single source defense update can protect
    independently specialized, held-out target policies
    without target-task adversarial fine-tuning.

    \item We identify a multi-source defense composition
    challenge and introduce SN-TRV, a source-norm-calibrated
    transfer rule that improves robustness over naive
    averaging without target-side adversarial optimization
    or additional inference modules.

    \item We systematically characterize defense reuse
    across fixed and adaptive attacks, source defense
    recipes, VLA backbones, and physical manipulation,
    revealing its effectiveness, attack-coverage
    dependence, and practical limitations.
\end{itemize}

% ================================================================
% II. RELATED WORK
% ================================================================
% ================================================================
% II. RELATED WORK
% ================================================================
\section{Related Work}

\subsection{Vision-Language-Action Policies and Robustness Evaluation}

Vision-Language-Action (VLA) policies such as RT-2, OpenVLA, Octo, and
$\pi_0$ have demonstrated strong transfer and generalization across robotic
manipulation tasks
~\cite{zitkovich2023rt2,kim2025openvla,octo2024,black2025pi0}.
LIBERO provides a widely used benchmark for evaluating knowledge transfer
and long-horizon manipulation~\cite{liu2023libero}.

Recent work has further examined VLA robustness under environmental and
adversarial variations. EVA-VLA and LIBERO-Plus evaluate robustness under
realistic distribution shifts~\cite{liu2025evavla,fei2026liberoplus},
while multi-view studies reveal failure modes such as view collapse under
adversarial training~\cite{waseda2026viewcollapse}.
These works mainly evaluate the robustness of a given policy after training
or adaptation; we instead ask whether learned robustness can be reused
across independently specialized VLA policies.

\subsection{Adversarial Attacks and Defenses for VLA Policies}

Physical adversarial perturbations can directly alter closed-loop robot
actions~\cite{wang2025adversarialvla}.
EDPA introduces model-agnostic adversarial patches and adversarial
fine-tuning for VLA policies~\cite{xu2025edpa}, while subsequent studies
consider universal transfer, attention manipulation, and broader attack
objectives
~\cite{lu2026uparfas,fu2026vlahijack,li2025attackvla}.
Recent cross-task attacks such as UniTexture further suggest that VLA
vulnerability can contain task-shared structure~\cite{dai2026unitexture}.

Correspondingly, SARF, VLA-Guard, RobustVLA, StableVLA, RoVLA, and
CertVLA improve robustness through adversarial fine-tuning, representation
correction, lightweight adaptation, or certification
~\cite{zhang2026sarf,yin2026vlaguard,guo2026robustvla,
fu2026stablevla,luo2026rovla,lu2026certvla}.
However, these defenses primarily optimize the policy being protected,
typically requiring target-specific robustness optimization.
Our work instead investigates whether defense knowledge learned on source
tasks can protect a held-out target policy without target-side adversarial
fine-tuning.

\subsection{Robustness Transfer and Parameter-Space Reuse}

Adversarial robustness transfer has been studied in
conventional vision, including robustness inheritance
from pretrained models to downstream tasks
\cite{yamada2022robustness,nern2023transfer}.
These studies mainly examine downstream adaptation,
rather than transferring defense-induced updates
between independently specialized robot policies.

Task Arithmetic~\cite{ilharco2023taskarithmetic}
and TIES-Merging~\cite{yadav2023ties} demonstrate
the reuse and combination of fine-tuning updates.
RETAIN~\cite{yadav2025retain} interpolates
pretrained and task-fine-tuned VLA weights
to preserve generalist capabilities and improve
out-of-distribution task generalization,
while VLA-Pro~\cite{si2026vlapro} transfers
procedural knowledge through adapters.
In contrast, we transfer source-learned adversarial
defense updates to independently specialized
target policies without target-side adversarial
fine-tuning.

MAGIC~\cite{li2025magic} studies magnitude
deviation in model merging and introduces
layer-wise calibration, including data-free
Weight Space Calibration (WSC).
Rather than claiming magnitude calibration
as a general novelty, we investigate source-only
global norm calibration for cross-task adversarial
defense reuse, with a tested WSC configuration
as a baseline.

% ================================================================
% III. METHOD
% ================================================================

\section{Methodology}

\begin{figure*}[!t]
    \centering
    \includegraphics[width=\textwidth]{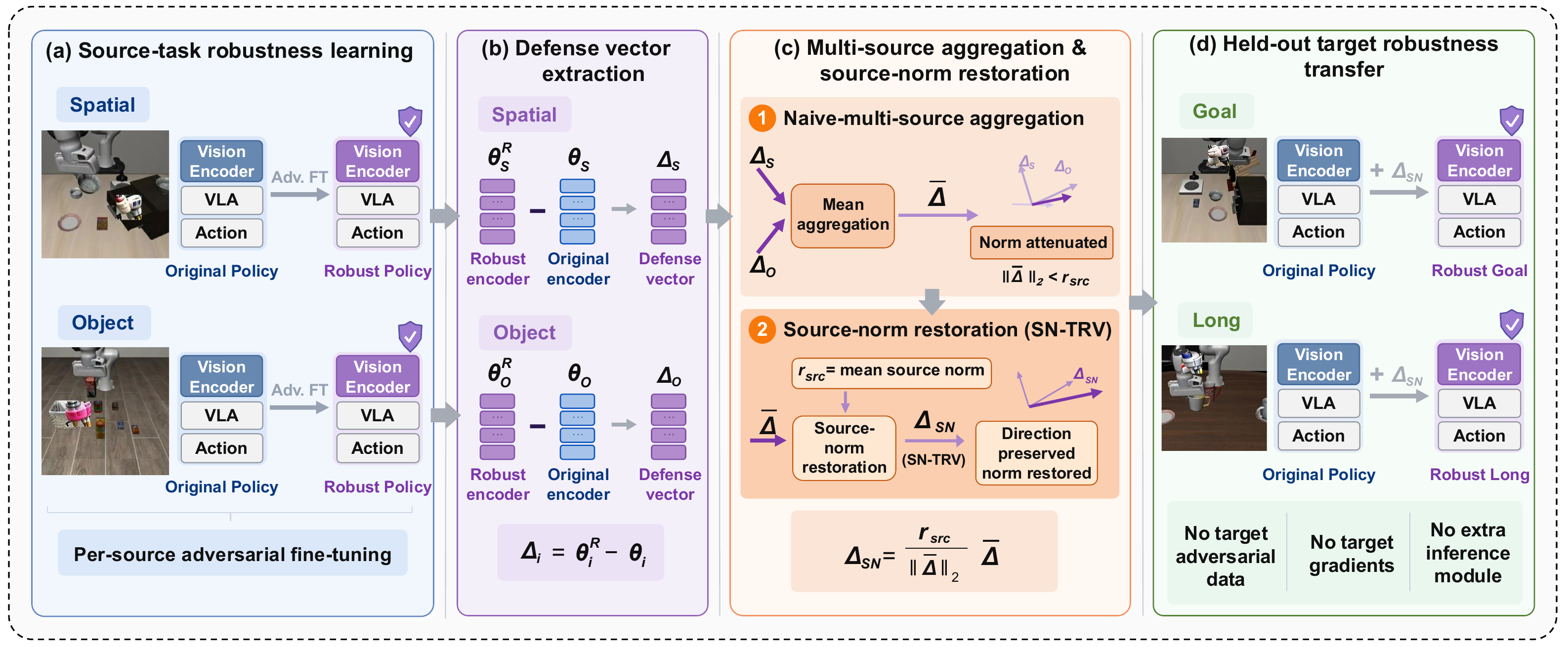}
    \caption{
    Cross-task defense reuse with SN-TRV.
(a) Source-task adversarial fine-tuning learns visual defense updates.
(b) Parameter differences are extracted from aligned source policies.
(c) A source-consensus direction and a source-calibrated magnitude are
computed separately and recomposed into SN-TRV.
(d) The resulting update is applied to a held-out target visual encoder
without target adversarial data, gradient adaptation, or additional
inference modules.
    }
    \label{fig:sntrv_overview}
\end{figure*}

\subsection{Cross-Task Defense Reuse Setting}

We transfer adversarial robustness from source policies to a
held-out target policy excluded from defense-vector construction.
Write a policy as $\Theta=(\theta,\psi)$, where $\theta$ contains
visual-encoder parameters and $\psi$ contains all remaining
components. For source $i$, the original and robust policies are
$\Theta_i=(\theta_i,\psi_i)$ and
$\Theta_i^R=(\theta_i^R,\psi_i)$.
The defense update is
\begin{equation}
    \Delta_i=\theta_i^R-\theta_i.
    \label{eq:defense_vector}
\end{equation}
Here, held-out means that the target policy does not
contribute to defense-vector construction or calibration;
it may already have been trained for its manipulation task.
Each $\Delta_i$ isolates the visual parameter change
induced by adversarial fine-tuning on source task $i$.
Unlike interpolation between a pretrained policy and its
target-task-finetuned counterpart, we reuse only this
source-learned defense increment and add it to the
independently specialized target encoder.
This operation requires access to target parameters but
does not involve target-side adversarial optimization.

Within each backbone, source and target visual encoders share
the same architecture, initialization, and parameterization,
enabling coordinate-aligned transfer.
We view each update as a vector of corresponding visual-encoder
parameters when computing norms and inner products.
Only the target visual encoder is modified; its action head,
proprioceptive modules, and other non-visual components remain
unchanged.
Directly adding $\Delta_i$ provides the single-source transfer
baseline. Fig.~\ref{fig:sntrv_overview} summarizes the
multi-source procedure.

\subsection{Multi-Source Defense Update Composition}

Given $M$ source defense updates, we estimate their aggregate
direction as
\begin{equation}
    \bar{\Delta}=\frac{1}{M}\sum_{i=1}^{M}\Delta_i,
    \qquad
    u_{\mathrm{def}}=
    \frac{\bar{\Delta}}{\|\bar{\Delta}\|_2},
    \label{eq:consensus_direction}
\end{equation}
assuming $\bar{\Delta}\neq 0$.
Averaging combines aligned components while allowing
disagreements to cancel.
Because raw updates are averaged, larger-norm sources exert
greater influence on the direction.

Define the mean source-update norm:
\begin{equation}
    r_{\mathrm{src}}=
    \frac{1}{M}\sum_{i=1}^{M}\|\Delta_i\|_2,
    \qquad
    \|\bar{\Delta}\|_2\leq r_{\mathrm{src}}.
    \label{eq:source_norm}
\end{equation}
The inequality follows from the triangle inequality.
Unlike $r_{\mathrm{src}}$, the aggregate norm depends on
cross-source alignment and can be attenuated by disagreement.
Thus, transferring $\bar{\Delta}$ couples direction estimation
with magnitude selection.
Such geometric attenuation does not establish that a
proportionally weaker update is preferable for target
robustness, motivating separate magnitude calibration.

\subsection{Source-Norm-Calibrated Defense Transfer}

SN-TRV retains the aggregate direction and assigns the mean
source norm:
\begin{equation}
    \Delta_{\mathrm{SN}}
    =r_{\mathrm{src}}u_{\mathrm{def}}
    =\frac{r_{\mathrm{src}}}{\|\bar{\Delta}\|_2}\bar{\Delta}.
    \label{eq:sn_trv}
\end{equation}
This preserves the direction and parameter support of
Mean-$\Delta$ while changing only its magnitude.

For two nonzero source updates, let $r_i=\|\Delta_i\|_2$
and $c=\cos(\Delta_1,\Delta_2)$.
Equation~\eqref{eq:sn_trv} gives
\[
\lambda_{\mathrm{SN}}
=
\frac{r_1+r_2}
{\sqrt{r_1^2+r_2^2+2r_1r_2c}}.
\]
For equal-norm orthogonal updates, this reduces to $\sqrt{2}$;
for collinear updates pointing in the same direction, it equals $1$.
The factor therefore depends on source norms and alignment rather
than only on the number of sources.
These special cases motivate the $\sqrt{2}$ comparison in
Section~\ref{sec:transfer_ablations}; equal norms and orthogonality
are illustrative conditions, not assumptions imposed on the learned
updates.

Equivalently, SN-TRV solves a source-norm-constrained
consensus problem:
\begin{equation}
    \Delta_{\mathrm{SN}}
    =
    \underset{\|\Delta\|_2=r_{\mathrm{src}}}{\arg\min}
    \sum_{i=1}^{M}\|\Delta-\Delta_i\|_2^2.
    \label{eq:constrained_consensus}
\end{equation}
At fixed norm, minimizing this objective is equivalent to
maximizing $\langle\Delta,\bar{\Delta}\rangle$, yielding
Eq.~\eqref{eq:sn_trv}.
This is a geometric characterization, not a guarantee of
optimal target robustness.

The transferred policy is
$\Theta_T^{\mathrm{SN}}=(\theta_T+\Delta_{\mathrm{SN}},\psi_T)$.
The reference magnitude is determined solely from source
updates, without target-task adversarial data or target-side
gradient optimization.
The resulting policy retains its original architecture and
inference pipeline.

% ================================================================
% IV. EXPERIMENTS
% ================================================================
\section{Experiments}

% ----------------------------------------------------------------
% A. Experimental Setup
% ----------------------------------------------------------------
\subsection{Experimental Setup}

\textbf{Benchmarks and policies.}
We evaluate on the four LIBERO suites: Spatial, Object, Goal,
and Long. OpenVLA-OFT is the main backbone, using primary-
and wrist-camera observations together with proprioceptive inputs.
Each suite contains 10 manipulation tasks. Unless otherwise
specified, we evaluate 50 rollouts per task, yielding 500 episodes
per suite-level condition, and report task success rate (SR, \%).
Spatial and Object are the default source suites; Goal and Long
are excluded from defense-vector construction.
Only visual-encoder parameters are transferred. Each target retains
its own action head, proprioceptive modules, normalization
statistics, tokenizer, and other non-visual components.

\textbf{Attacks and source defenses.}
EDPA~\cite{xu2025edpa} is the primary threat.
On OpenVLA-OFT, Primary and Wrist attack the corresponding
camera individually, while Dual attacks both cameras simultaneously.
Unless specified as adaptive, patches are optimized against the
original target policy.
Source robust policies are obtained through EDPA adversarial
fine-tuning (EDPA-AF), which updates the visual encoder for
50k steps. Target-specific EDPA-AF is a robustness reference,
not a budget-matched transfer baseline, because it uses target
adversarial data and target-side optimization.
We additionally evaluate SARF-derived updates
and Attention-Guided Semantic Disruption
(AGSD) attacks~\cite{zhang2026sarf},
as well as transfer within OpenVLA and $\pi_0$.

\textbf{Transfer baselines.}
We compare Original, direct Spatial and Object defense-vector
transfer, the arithmetic mean of both updates (Mean-$\Delta$),
Cos-Select, and SN-TRV.
Cos-Select partitions the visual encoder into 53 parameter groups,
including Transformer blocks and grouped embedding/output
parameters. After excluding two zero-update groups, it ranks
the remaining 51 groups by the signed cosine similarity between
Spatial and Object updates. The top 50\% ($K=26$) retain the
equally averaged source updates; all other updates are zeroed.
The same source-only mask is used for all targets without norm
compensation. No target checkpoint, evaluation result, or
adversarial data is used to select this mask.

Cos-Select+SN retains this mask and rescales the selected update
as $\Delta_{\mathrm{Cos+SN}}=
(r_{\mathrm{src}}/\|\Delta_{\mathrm{Cos}}\|_2)
\Delta_{\mathrm{Cos}}$.
It therefore matches the global norm of SN-TRV while retaining
the direction and parameter support of Cos-Select.
We also evaluate the weight-space calibration baseline (WSC)
related to MAGIC~\cite{li2025magic}; comparisons refer to the
tested WSC configuration.
% AUTHOR CHECK: The supplied manuscript does not specify the WSC
% calibration formula, grouping, sensitivity masking, or data use.
% Do not invent these details or identify this baseline as the full
% MAGIC framework. Record the missing specification in the report.

% ----------------------------------------------------------------
% B. Cross-Task Robustness Transfer
% ----------------------------------------------------------------
\subsection{Cross-Task Robustness Transfer}

We first test whether adversarial robustness
learned on a single source suite can be reused
by an independently specialized, held-out
target policy, without multi-source aggregation.
Table~\ref{tab:main_results} compares clean
performance and robustness under single-
and dual-camera EDPA.

\begin{table}[!t]
\centering
\caption{
Cross-task robustness transfer on held-out
OpenVLA-OFT policies under EDPA.
EDPA-AF uses target-task adversarial fine-tuning
and serves as a reference rather than a
source-only transfer baseline.
All values are success rates (\%).
}
\label{tab:main_results}

\tablesetup

\begin{tabular*}{\columnwidth}{
@{\extracolsep{\fill}}
llcccc
@{}
}
\toprule
Target & Method & Clean & Primary & Wrist & Dual \\
\midrule

\multirow{7}{*}{Goal}
& Original
& \textbf{97.2} & 88.8 & 61.4 & 22.2 \\

& Spatial $\Delta$
& 93.0 & 90.0 & 92.4 & 87.2 \\

& Object $\Delta$
& 92.8 & 90.4 & 86.0 & 80.0 \\

& Mean-$\Delta$
& 94.6 & 84.4 & 81.4 & 76.6 \\

& Cos-Select
& 94.8 & \underline{92.0} & 87.0 & 80.8 \\

& EDPA-AF~\cite{xu2025edpa}
& \underline{96.4}
& \textbf{97.0}
& \underline{96.2}
& \textbf{96.2} \\

& \textbf{SN-TRV}
& \textbf{97.2}
& \textbf{97.0}
& \textbf{97.2}
& \underline{95.4} \\

\midrule

\multirow{7}{*}{Long}
& Original
& \textbf{93.2} & 68.4 & 48.2 & 10.0 \\

& Spatial $\Delta$
& 90.2 & 79.8 & 81.6 & 68.6 \\

& Object $\Delta$
& \underline{92.6} & 78.4 & 79.0 & 60.6 \\

& Mean-$\Delta$
& 90.4 & 76.8 & 76.4 & 62.0 \\

& Cos-Select
& 90.8 & 77.4 & 76.2 & 58.2 \\

& EDPA-AF~\cite{xu2025edpa}
& \underline{92.6}
& \textbf{86.8}
& \textbf{90.8}
& \textbf{85.2} \\

& \textbf{SN-TRV}
& 92.4
& \underline{86.2}
& \underline{87.6}
& \underline{79.4} \\

\bottomrule
\end{tabular*}
\end{table}

Under Dual EDPA, the original policies achieve
only $22.2\%$ and $10.0\%$ success on Goal and Long.
Directly transferring the Spatial defense update
raises success to $87.2\%$ and $68.6\%$,
corresponding to gains of $65.0$ and $58.6$
percentage points.
Object transfer also achieves substantial
improvements, reaching $80.0\%$ and $60.6\%$.
These single-source gains demonstrate that
source-learned adversarial robustness can be
reused across manipulation suites independently
of any multi-source merging or target-task
adversarial fine-tuning.

\begin{figure}[!t]
    \centering
    \includegraphics[width=\linewidth]{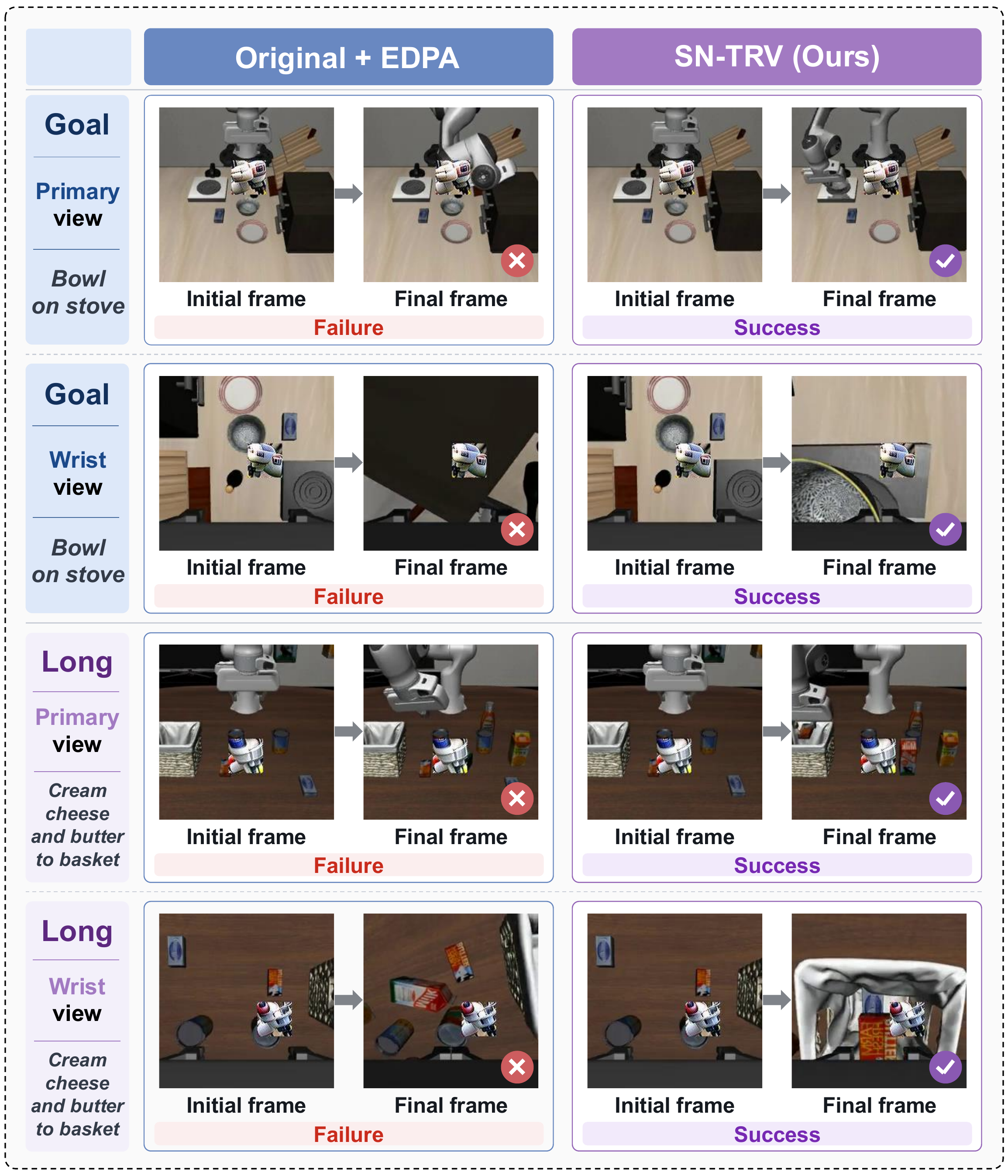}
    \caption{
    Simulation rollouts under EDPA on held-out Goal and Long tasks.
    With matched initial conditions and perturbations, the original
    policies fail, whereas SN-TRV restores successful manipulation
    without target-task adversarial fine-tuning.
    }
    \label{fig:sim_qualitative}
\end{figure}

Reusing multiple source defenses introduces a
separate challenge.
Naive Mean-$\Delta$ achieves only $76.6\%$
and $62.0\%$ Dual success, below direct
Spatial transfer on both targets.
SN-TRV instead reaches $95.4\%$ and $79.4\%$,
improving over Mean-$\Delta$ by $18.8$ and
$17.4$ percentage points, and over Spatial
transfer by $8.2$ and $10.8$ points.
Its clean success remains $97.2\%$ and
$92.4\%$, compared with $97.2\%$ and $93.2\%$
for the original policies.
These results separate the existence of
transferable robustness from the additional
benefit of calibrated multi-source reuse.
Fig.~\ref{fig:sim_qualitative} illustrates
representative closed-loop outcomes.

The single-camera conditions show why Dual robustness must be
evaluated separately.
On Long, Original achieves $68.4\%$ under Primary and $48.2\%$
under Wrist, but only $10.0\%$ under Dual.
SN-TRV reaches $86.2\%$, $87.6\%$, and $79.4\%$, respectively.
Its improvement therefore extends across both individual-camera
attacks and simultaneous perturbations.
The remaining gap between single- and dual-camera success also
shows that strong performance with only one attacked view does
not fully characterize performance when both visual inputs are
perturbed.

These results also illustrate the target-side optimization benefit.
Averaged over Goal and Long, SN-TRV achieves $87.4\%$ Dual
success without target-side adversarial fine-tuning, compared
with $90.7\%$ for target-specific EDPA-AF, which requires
50k defense optimization steps per target.
This comparison assumes that the source robust policies are
already available and excludes their initial training cost.

% ----------------------------------------------------------------
% C. Direction--Magnitude Analysis
% ----------------------------------------------------------------
\subsection{Direction--Magnitude Analysis of Defense Transfer}

\begin{figure}[!t]
    \centering
    \includegraphics[width=0.94\linewidth]{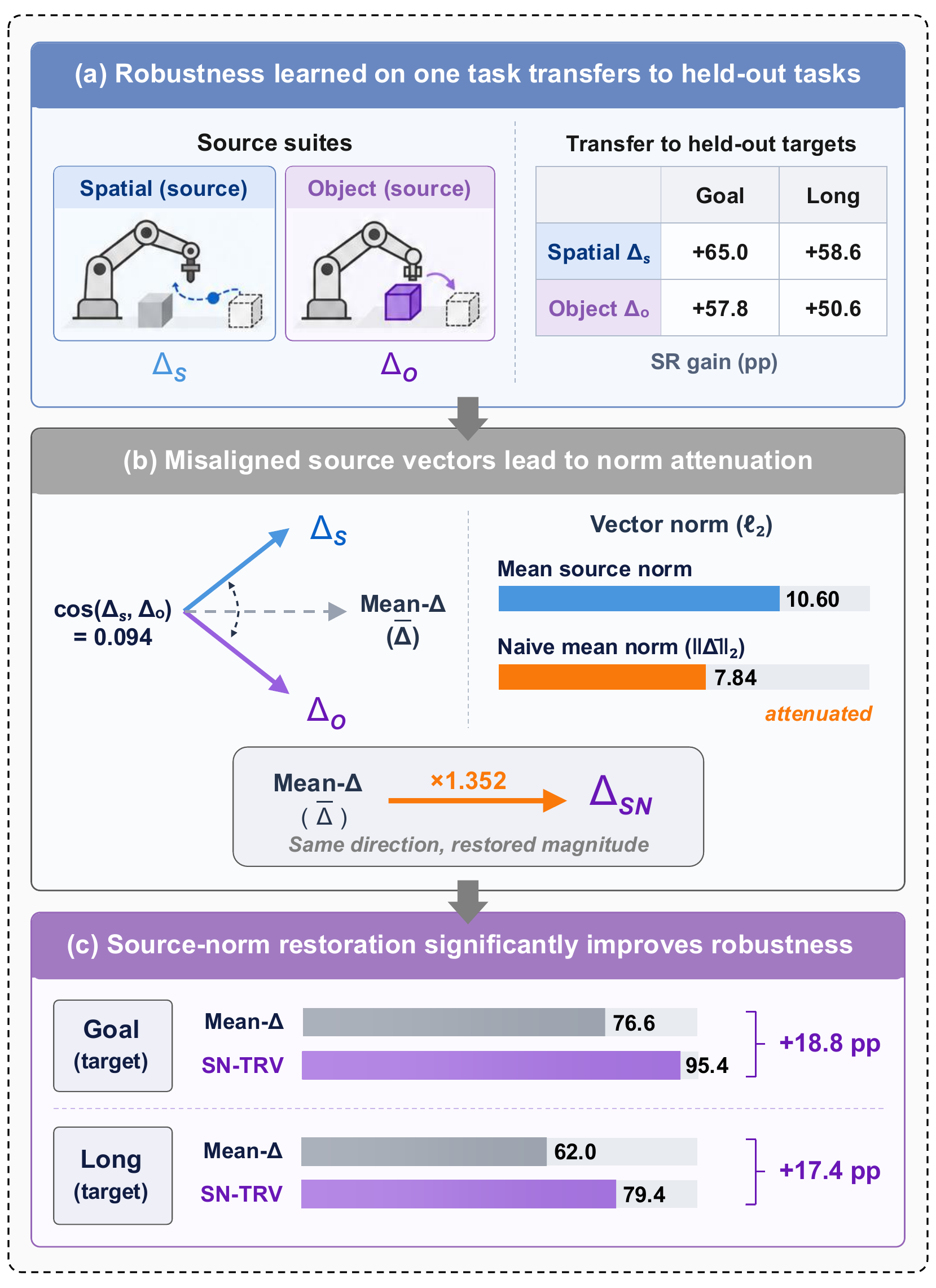}
    \caption{
Cross-task defense reuse and source-update geometry.
(a) Single-source gains from Table~\ref{tab:main_results}.
(b) Weakly aligned source updates yield an attenuated mean norm.
(c) Source-norm restoration improves fixed dual-camera EDPA
success while preserving the mean update direction.
}
    \label{fig:cross_task_map}
\end{figure}

We next examine why combining individually
transferable source defenses can underperform
direct single-source transfer.
As shown in Fig.~\ref{fig:cross_task_map}, Spatial and Object
updates have a global cosine similarity of $0.0942$.
Their arithmetic mean has an $\ell_2$ norm of $7.841$, below
the mean source-update norm of $10.600$, consistent with
Eq.~\eqref{eq:source_norm}.
These measurements establish weak global alignment and
aggregate-norm attenuation, but do not by themselves establish
a corresponding loss of target robustness.

Mean-$\Delta$ and SN-TRV provide a controlled comparison:
their update direction and parameter support are identical,
while SN-TRV restores the norm to $10.600$, scaling the mean
by approximately $1.352$.
Fig.~\ref{fig:cross_task_map}(c) reports the resulting fixed
Dual EDPA gains of $18.8$ and $17.4$ percentage points on
Goal and Long.
This comparison demonstrates an empirical benefit of increasing
magnitude along the same direction, rather than proving an
optimal target update norm.
Alternative magnitudes and update constructions are examined
in Section~\ref{sec:transfer_ablations}.

% ----------------------------------------------------------------
% D. Adaptive Attacks
% ----------------------------------------------------------------
\subsection{Robustness under Adaptive Attacks}

Fixed-patch gains could partly reflect attack--model mismatch:
the transferred policy differs from the policy used to optimize
the patch. We therefore re-optimize EDPA against each evaluated
victim using the same 50k-step attack budget.
Primary- and wrist-camera patches are optimized independently
against the corresponding policy and then applied simultaneously.

\begin{table}[!t]
\centering
\caption{
Robustness under fixed and adaptively re-optimized dual-camera
EDPA on OpenVLA-OFT. Fixed and Adaptive are success rates
(\%); Drop is Fixed$-$Adaptive in percentage points.
}
\label{tab:adaptive_edpa}

\tablesetup

\begin{tabular*}{\columnwidth}{
@{\extracolsep{\fill}}
llccc
@{}
}
\toprule
Target & Method & Fixed & Adaptive & Drop \\
\midrule

\multirow{5}{*}{Goal}
& Original
& 22.2 & 22.2 & 0.0 \\

& Spatial $\Delta$
& 87.2 & 72.6 & 14.6 \\

& Mean-$\Delta$
& 76.6 & 61.4 & 15.2 \\

& EDPA-AF~\cite{xu2025edpa}
& \textbf{96.2} & \textbf{85.4} & 10.8 \\

& \textbf{SN-TRV}
& \underline{95.4} & \underline{81.2} & 14.2 \\

\midrule

\multirow{5}{*}{Long}
& Original
& 10.0 & 10.0 & 0.0 \\

& Spatial $\Delta$
& 68.6 & 34.6 & 34.0 \\

& Mean-$\Delta$
& 62.0 & 28.6 & 33.4 \\

& EDPA-AF~\cite{xu2025edpa}
& \textbf{85.2} & \underline{43.8} & 41.4 \\

& \textbf{SN-TRV}
& \underline{79.4} & \textbf{46.6} & 32.8 \\

\bottomrule
\end{tabular*}
\end{table}

Table~\ref{tab:adaptive_edpa} shows that SN-TRV retains a
substantial advantage over Mean-$\Delta$ after re-optimization.
On Goal, adaptive success increases from $61.4\%$ to $81.2\%$;
on Long, it increases from $28.6\%$ to $46.6\%$.
The gains of $19.8$ and $18.0$ percentage points are comparable
to the fixed-attack gains reported above.
Thus, the benefit of source-norm calibration persists under
this adaptive attack protocol.
The main advantage is higher absolute success, rather than a
substantially smaller fixed-to-adaptive decrease.

SN-TRV also exceeds direct Spatial transfer,
improving adaptive Dual success by $8.6$ and
$12.0$ percentage points on Goal and Long.
Its slightly higher success than EDPA-AF on Long ($46.6\%$
versus $43.8\%$) does not establish general superiority over
target-specific adversarial fine-tuning.

% ----------------------------------------------------------------
% E. Source Defense and Attack Coverage
% ----------------------------------------------------------------
\subsection{Defense-Recipe and Attack-Scope Analysis}

We next ask whether the source defense determines which attacks
are resisted after transfer.
Table~\ref{tab:defense_attack} evaluates EDPA-AF- and
SARF-derived updates against EDPA and AGSD on OpenVLA,
where matched source checkpoints are available under the same
architecture and evaluation protocol.
Spatial and Object remain the sources, with Goal and Long
excluded from vector construction.

\begin{table}[!t]
\centering
\caption{
Defense-recipe and attack-scope analysis on OpenVLA.
EDPA-AF and SARF are defense training methods,
while SN-TRV transfers source-learned defense
updates. AF and SARF in parentheses indicate
the source defense recipes.
All values are success rates (\%).
}
\label{tab:defense_attack}

\tablesetup

\begin{tabular*}{\columnwidth}{
@{\extracolsep{\fill}}
llccc
@{}
}
\toprule
Target & Method & Clean & EDPA & AGSD \\
\midrule

\multirow{7}{*}{Goal}
& Original
& \textbf{79.2} & 0.0 & 0.0 \\

& EDPA-AF~\cite{xu2025edpa}
& 77.2 & 60.4 & 2.8 \\

& SARF~\cite{zhang2026sarf}
& \underline{79.0}
& \textbf{74.6}
& \textbf{73.0} \\

& Mean-$\Delta$ (AF)
& 75.8 & 46.2 & 1.4 \\

& \textbf{SN-TRV (AF)}
& 74.4 & 59.2 & 2.0 \\

& Mean-$\Delta$ (SARF)
& 74.0 & 51.2 & 52.4 \\

& \textbf{SN-TRV (SARF)}
& 76.2
& \underline{72.4}
& \underline{71.8} \\

\midrule

\multirow{7}{*}{Long}
& Original
& \textbf{53.8} & 0.0 & 0.0 \\

& EDPA-AF~\cite{xu2025edpa}
& 51.0 & 27.6 & 1.0 \\

& SARF~\cite{zhang2026sarf}
& \underline{53.4}
& \textbf{46.0}
& \textbf{43.2} \\

& Mean-$\Delta$ (AF)
& 51.2 & 12.8 & 0.4 \\

& \textbf{SN-TRV (AF)}
& 52.2 & 27.4 & 0.8 \\

& Mean-$\Delta$ (SARF)
& 50.4 & 33.2 & \underline{31.2} \\

& \textbf{SN-TRV (SARF)}
& 52.0
& \underline{44.0}
& \textbf{43.2} \\

\bottomrule
\end{tabular*}
\end{table}

With EDPA-AF sources, SN-TRV achieves $59.2\%$ and $27.4\%$
EDPA success on Goal and Long, but only $2.0\%$ and $0.8\%$
under AGSD.
With SARF sources, its EDPA success reaches $72.4\%$ and
$44.0\%$, while AGSD success reaches $71.8\%$ and $43.2\%$.
For each source recipe, SN-TRV also improves over the
corresponding Mean-$\Delta$ on both attacks and targets.

Comparison with the corresponding defense reference gives a
complementary measure of transfer.
Under EDPA, SN-TRV with AF sources is $1.2$ and $0.2$ percentage
points below EDPA-AF on Goal and Long.
With SARF sources, the EDPA gaps to SARF are $2.2$ and $2.0$
points, while the AGSD gaps are $1.2$ and $0.0$ points.
These point estimates indicate proximity to the corresponding
defense reference, not statistical equivalence.
Together with the attack-specific results, they show that the
transferred updates retain a similar coverage pattern rather than
creating a new attack-agnostic defense.

These comparisons distinguish the source defense from the
transfer rule. The source recipe strongly shapes transferred
attack coverage, while SN-TRV governs update aggregation
and magnitude calibration.
In particular, calibrating EDPA-AF updates does not make them
effective against AGSD in these evaluations; broader protection
is observed when the source updates themselves come from SARF.

% ----------------------------------------------------------------
% F. Generalization across VLA Architectures
% ----------------------------------------------------------------
\subsection{Generalization across VLA Architectures}

We examine whether cross-task transfer is confined to
OpenVLA-OFT. Table~\ref{tab:architecture} additionally
evaluates OpenVLA and $\pi_0$.
Source updates are constructed and transferred separately within
each backbone; parameters are never transferred between
incompatible architectures.

\begin{table}[!t]
\centering
\caption{
Cross-task robustness transfer across VLA architectures on held-out
Goal and Long suites under EDPA. All values are success rates (\%).
}
\label{tab:architecture}

\tablesetup

\begin{tabular*}{\columnwidth}{
@{\extracolsep{\fill}}
lllcc
@{}
}
\toprule
Backbone & Target & Method & Clean & EDPA \\
\midrule

\multirow{8}{*}{OpenVLA}

& \multirow{4}{*}{Goal}
& Original & \textbf{79.2} & 0.0 \\

&
& Spatial $\Delta$ & 70.4 & 57.0 \\

&
& EDPA-AF~\cite{xu2025edpa}
& \underline{77.2} & \textbf{60.4} \\

&
& \textbf{SN-TRV}
& 74.4 & \underline{59.2} \\

\cmidrule(lr){2-5}

& \multirow{4}{*}{Long}
& Original & \textbf{53.8} & 0.0 \\

&
& Spatial $\Delta$ & 46.2 & 12.6 \\

&
& EDPA-AF~\cite{xu2025edpa}
& 51.0 & \textbf{27.6} \\

&
& \textbf{SN-TRV}
& \underline{52.2} & \underline{27.4} \\

\midrule

\multirow{8}{*}{OpenVLA-OFT}

& \multirow{4}{*}{Goal}
& Original & \textbf{97.2} & 22.2 \\

&
& Spatial $\Delta$ & 93.0 & 87.2 \\

&
& EDPA-AF~\cite{xu2025edpa}
& \underline{96.4} & \textbf{96.2} \\

&
& \textbf{SN-TRV}
& \textbf{97.2} & \underline{95.4} \\

\cmidrule(lr){2-5}

& \multirow{4}{*}{Long}
& Original & \textbf{93.2} & 10.0 \\

&
& Spatial $\Delta$ & 90.2 & 68.6 \\

&
& EDPA-AF~\cite{xu2025edpa}
& \underline{92.6} & \textbf{85.2} \\

&
& \textbf{SN-TRV}
& 92.4 & \underline{79.4} \\

\midrule

\multirow{8}{*}{$\pi_0$}

& \multirow{4}{*}{Goal}
& Original & \textbf{93.8} & 49.6 \\

&
& Spatial $\Delta$ & 87.6 & 85.6 \\

&
& EDPA-AF~\cite{xu2025edpa}
& 92.2 & \underline{89.4} \\

&
& \textbf{SN-TRV}
& \underline{92.8} & \textbf{89.8} \\

\cmidrule(lr){2-5}

& \multirow{4}{*}{Long}
& Original & \textbf{81.0} & 32.8 \\

&
& Spatial $\Delta$ & 71.0 & 63.6 \\

&
& EDPA-AF~\cite{xu2025edpa}
& \underline{78.6} & \textbf{72.6} \\

&
& \textbf{SN-TRV}
& 75.8 & \underline{71.4} \\

\bottomrule
\end{tabular*}
\end{table}

On OpenVLA, SN-TRV raises EDPA success from $0.0\%$ to
$59.2\%$ on Goal and $27.4\%$ on Long, exceeding the reported
Spatial transfer on both targets.
Clean success changes from $79.2\%$ to $74.4\%$ on Goal
and from $53.8\%$ to $52.2\%$ on Long.
The OpenVLA-OFT entries are repeated from
Table~\ref{tab:main_results} for reference.

For $\pi_0$, source updates are obtained by adversarially
fine-tuning the same clean checkpoint separately on Spatial
and Object, excluding Goal and Long from vector construction.
The EDPA-AF reference is a four-suite model with target-suite
exposure and is therefore not a target-free baseline.
SN-TRV increases EDPA success from $49.6\%$ to $89.8\%$
on Goal and from $32.8\%$ to $71.4\%$ on Long.
However, Long clean success decreases from $81.0\%$ to
$75.8\%$.
These within-backbone comparisons support transfer beyond
OpenVLA-OFT, while showing that its clean-performance cost
remains architecture- and task-dependent.

% ----------------------------------------------------------------
% G. Magnitude Sensitivity and Transfer Ablations
% ----------------------------------------------------------------
\subsection{Magnitude Sensitivity and Transfer Ablations}
\label{sec:transfer_ablations}

\textbf{Magnitude sensitivity.}
The geometry analysis compares only Mean-$\Delta$ and SN-TRV.
To examine other magnitudes, we evaluate
$\theta_T^{(\lambda)}=\theta_T+\lambda\bar{\Delta}$ for
$\lambda\in\{1.0,1.25,\lambda_{\mathrm{SN}},\sqrt{2},2.0\}$,
keeping the source updates, target policies, and fixed-attack
protocol unchanged.
The SN-TRV coefficient remains
$\lambda_{\mathrm{SN}}=r_{\mathrm{src}}/\|\bar{\Delta}\|_2$;
it is not selected using the target success rates in this sweep.
For two sources, $\lambda=2$ corresponds to summing
the source updates:
$2\bar{\Delta}=\Delta_{\mathrm{Spatial}}+\Delta_{\mathrm{Object}}$.

\begin{figure}[!t]
    \centering
    \includegraphics[width=\columnwidth]
    {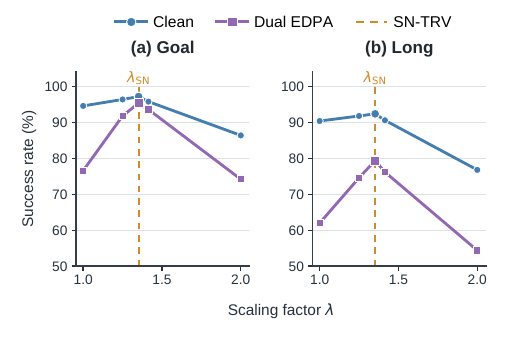}
    \caption{
Sensitivity to update magnitude on OpenVLA-OFT.
(a) Goal and (b) Long success rates under clean inputs and
fixed dual-camera EDPA.
All variants apply $\lambda\bar{\Delta}$ to the same target
visual encoder. Markers denote evaluated scales; the dashed
line marks the source-derived $\lambda_{\mathrm{SN}}$.
}
    \label{fig:magnitude_sensitivity}
\end{figure}

\begin{table}[!t]
\centering
\caption{
Comparison of source-side defense update constructions on
held-out OpenVLA-OFT policies. Dual denotes fixed dual-camera
EDPA. All values are success rates (\%).
}
\label{tab:transfer_ablation}

\tablesetup

\begin{tabular*}{\columnwidth}{
@{\extracolsep{\fill}}
lcccc
@{}
}
\toprule

& \multicolumn{2}{c}{Goal}
& \multicolumn{2}{c}{Long} \\

\cmidrule(lr){2-3}
\cmidrule(lr){4-5}

Method & Clean & Dual & Clean & Dual \\
\midrule

Mean-$\Delta$
& 94.6 & 76.6 & 90.4 & 62.0 \\

Cos-Select
& 94.8 & 80.8 & 90.8 & 58.2 \\

Cos-Select+SN
& \underline{96.4}
& \underline{89.4}
& \underline{92.0}
& \underline{70.6} \\

WSC~\cite{li2025magic}
& 95.6 & 84.2 & 91.4 & 68.8 \\

\midrule

\textbf{SN-TRV}
& \textbf{97.2}
& \textbf{95.4}
& \textbf{92.4}
& \textbf{79.4} \\

\bottomrule
\end{tabular*}
\end{table}

\begin{figure}[!t]
    \centering
    \includegraphics[width=\linewidth]{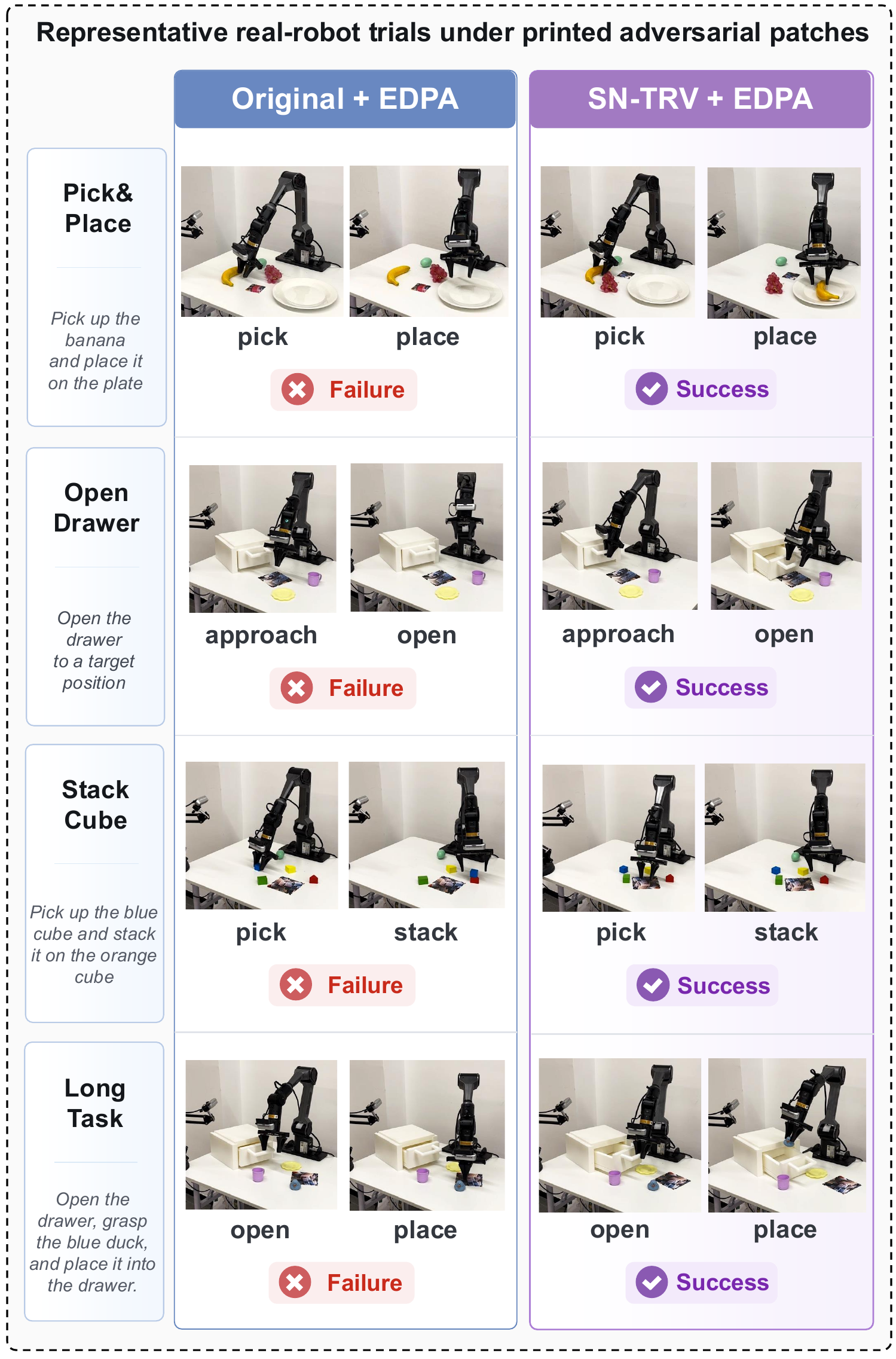}
    \caption{
    Representative PiPER trials under a printed EDPA patch.
    SN-TRV transfers robustness from other manipulation tasks to restore
    execution on held-out physical tasks without target-side adversarial
    fine-tuning.
    }
    \label{fig:real_robot}
\end{figure}

Fig.~\ref{fig:magnitude_sensitivity} shows that SN-TRV
($\lambda_{\mathrm{SN}}\approx1.352$) has the highest reported
clean and fixed Dual success among the five evaluated scales
on both targets.
The nearby scales are also effective: $\lambda=1.25$ reaches
$91.8\%$/$74.6\%$ Dual success on Goal/Long, and
$\lambda=\sqrt{2}$ reaches $93.6\%$/$76.2\%$, compared with
$95.4\%$/$79.4\%$ for SN-TRV.
At $\lambda=2$, Dual success falls to $74.2\%$/$54.4\%$
and clean success to $86.4\%$/$76.8\%$.
Thus, arbitrary amplification does not ensure better robustness.
The sweep supports a source-derived magnitude reference in this
setting, not a global optimum over continuous scales or a
statistically established advantage over nearby values.

\textbf{Update construction.}
We next compare alternative update constructions in
Table~\ref{tab:transfer_ablation}.
In particular, Cos-Select+SN and SN-TRV have the same global
update norm, allowing us to examine whether their different
update constructions still yield different transfer performance.
WSC provides an additional calibration baseline.

The original Cos-Select update has norm $5.477$.
Rescaling it to $r_{\mathrm{src}}=10.600$ improves Dual success
from $80.8\%$ to $89.4\%$ on Goal and from $58.2\%$ to
$70.6\%$ on Long.
Nevertheless, SN-TRV exceeds Cos-Select+SN by $6.0$ and
$8.8$ percentage points at the same global norm.

Clean performance also improves after rescaling Cos-Select:
from $94.8\%$ to $96.4\%$ on Goal and from $90.8\%$ to $92.0\%$
on Long.
SN-TRV further reaches $97.2\%$ and $92.4\%$, respectively.
Thus, in these reported results, the robustness improvements are
accompanied by higher clean success rather than a measured loss
of nominal performance.

Norm matching therefore improves the selected update, but does
not remove the performance difference between the tested
constructions. This comparison controls total norm, not direction,
parameter support, or layerwise magnitude allocation.
The tested WSC baseline reaches $84.2\%$ and $68.8\%$ Dual
success, below SN-TRV on both targets.
These results favor the complete mean update with source-norm
calibration among the evaluated constructions, without implying
superiority over every configuration of related merging methods.

% ----------------------------------------------------------------
% H. Physical-Robot Robustness Transfer
% ----------------------------------------------------------------

\subsection{Physical-Robot Robustness Transfer}

We evaluate $\pi_0$ policies
with dual-camera observations on a PiPER tabletop manipulator.
The evaluation covers four tasks: Pick-and-Place, Open Drawer,
Stack Cube, and Long Task, under printed EDPA patches with
randomized object poses, camera viewpoints, and sensing
conditions. Each task--condition pair is evaluated with
100 physical trials.

We adopt a held-out-task transfer protocol: each target
receives EDPA-AF defense updates from two other manipulation
tasks, without using target-task adversarial data for vector
construction. Single-$\Delta$ reports the mean success rate
of two independently transferred single-source updates,
rather than a merged policy. Fig.~\ref{fig:real_robot}
illustrates representative trials, while
Table~\ref{tab:real_robot} presents the quantitative results.

Under printed EDPA, the original policies' average success
rate decreases from $68.0\%$ under clean conditions to
$20.3\%$. SN-TRV raises attacked success to $45.8\%$,
compared with $27.5\%$ for Single-$\Delta$ and $48.5\%$
for target-specific EDPA-AF. This corresponds to gains of
$25.5$ and $18.3$ percentage points over the attacked
original policies and Single-$\Delta$, respectively.
The improvement over Single-$\Delta$ holds across all
four tasks, including Long Task ($40.0\%$ versus $22.0\%$).
Although SN-TRV remains below target-specific EDPA-AF,
it reuses defense updates learned on other tasks without
target-side adversarial fine-tuning.

\begin{table}[!t]
\centering
\caption{
Real-robot success rates (\%) on four PiPER manipulation
tasks under clean and printed EDPA conditions.
Each target uses defense updates from two other tasks.
Single-$\Delta$ averages two independently transferred
single-source defenses.
}
\label{tab:real_robot}

\tablesetup

\begin{tabular*}{\columnwidth}{
@{\extracolsep{\fill}}lccccc@{}
}
\toprule
& Clean & \multicolumn{4}{c}{Printed EDPA} \\
\cmidrule(lr){3-6}
Task & Original & Original
& Single-$\Delta$ & EDPA-AF & \textbf{SN-TRV} \\
\midrule

Pick \& Place
& 82.0 & 29.0 & 35.0
& \textbf{59.0} & \underline{55.0} \\

Open Drawer
& 70.0 & 22.0 & 28.0
& \textbf{50.0} & \underline{47.0} \\

Stack Cube
& 64.0 & 18.0 & 25.0
& \textbf{43.0} & \underline{41.0} \\

Long Task
& 56.0 & 12.0 & 22.0
& \textbf{42.0} & \underline{40.0} \\

\midrule

Average
& 68.0 & 20.3 & 27.5
& \textbf{48.5} & \underline{45.8} \\

\bottomrule
\end{tabular*}
\end{table}

\section{Conclusion}

We investigated cross-task adversarial defense
reuse in independently specialized VLA policies.
Our results show that even a single source
defense update can substantially improve
robustness on held-out manipulation suites
without target-task adversarial fine-tuning.
To better combine multiple source defenses,
SN-TRV applies source-only norm calibration
and improves robustness over naive aggregation
without additional inference modules.

Magnitude and update-construction analyses characterize
the benefits and limitations of this calibration, while
adaptive attacks confirm gains over naive averaging
under the evaluated threat protocol. Experiments across
VLA backbones and physical manipulation tasks further
support defense reuse. Its applicability nevertheless
depends on compatible parameter spaces and the attack
coverage learned by source defenses, while clean-performance
costs remain task- and architecture-dependent.

\bibliographystyle{IEEEtran}
\IEEEtriggeratref{0}
\bibliography{refs}

\end{document}